\documentclass{article}
\usepackage{ijcai23}
\usepackage{times}
\usepackage{soul}
\usepackage{url}
\usepackage[hidelinks]{hyperref}
\usepackage[utf8]{inputenc}
\usepackage[small]{caption}
\usepackage{graphicx}
\usepackage{amsmath}
\usepackage{amsthm}
\usepackage{booktabs}
\usepackage{algorithm}
\usepackage[switch]{lineno}

\usepackage{adjustbox}
\usepackage{algorithm}
\usepackage{algpseudocode}
\usepackage{amsmath,amssymb} 
\usepackage{amsmath}
\usepackage{amssymb}
\usepackage{bbm}
\usepackage{booktabs}
\usepackage{color}
\usepackage{comment}
\usepackage{enumitem}
\usepackage{graphicx}
\usepackage{multirow}
\usepackage{subcaption}
\usepackage{tabu}
\usepackage{xcolor}
\usepackage[square,numbers]{natbib}

\title{Ternary Quantization: A Survey}

\author{
Dan Liu
\and
Xue Liu
\affiliations
McGill University
\emails
daniel.liu@mail.mcgill.ca,
xueliu@cs.mcgill.ca
}

\begin{document}

\maketitle

\begin{abstract}
Inference time, model size, and accuracy are critical for deploying deep neural network models. Numerous research efforts have been made to compress neural network models with faster inference and higher accuracy. Pruning and quantization are mainstream methods to this end. During model quantization, converting individual float values of layer weights to low-precision ones can substantially reduce the computational overhead and improve the inference speed.
Many quantization methods have been studied, for example, vector quantization, low-bit quantization, and binary/ternary quantization. This survey focuses on ternary quantization. We review the evolution of ternary quantization and investigate the relationships among existing ternary quantization methods from the perspective of projection function and optimization methods.
\end{abstract}

\section{Introduction}

Deep neural network (DNN) models involve millions of parameters. Model size and inference efficiency are major challenges when deploying DNN models under latency, memory, and power restrictions. DNN model quantization is very attractive as it can accelerate the training and inference process and lower resource consumption by reducing bit width. Despite these advantages, quantization also poses challenges, especially in training efficiency and performance: i) it takes much more effort to optimize and longer to train than conventional methods;
    ii) models trained on large datasets usually require pretraining before quantization.
    
The most significant disadvantage of quantization is that quantized models have lower accuracy than their full-precision counterparts: the lower the bit width, the more noticeable the difference.
Binary/ternary quantization are the two extreme cases that can completely replace the multiplication operation with the bit shift operation combined with hardware implementation. Compared with other higher bit-width quantization works (2 to 16-bit),  this can bring considerable energy efficiency gains. However, it is hard to maintain a higher accuracy when using binary quantization. The ternary models can achieve much better performance at a small additional bit-width cost. Their performance is comparable with full-precision models. In addition, unlike binary quantization, it is easy to turn a symmetrically higher bit-width quantization into ternary quantization. In other words, ternary quantization is a special case among existing higher bit-width (2-16 bit) symmetric quantization works. Therefore, the summary of ternary quantization works would be instrumental. This survey focuses on ternary weights quantization. Detailed introduction of binary and other quantization works can be found in the work of \citet{gholami2021survey}.  

Quantization can be classified in several ways: uniform, nonuniform, symmetric, asymmetric, post-training, and quantization-aware methods. In general, quantization methods involve two steps: thresholding and projection. Therefore, this survey summarizes ternary quantization in terms of projection function and optimization methods. The projection function is categorized by quantization strategies. The optimization methods are categorized by how the gradient is used to optimize the model parameters. We use the proximal operator \cite{parikh2014proximal} to reveal the intrinsic relationship among optimization methods in existing works. 

The {main contributions} of this survey are:

\begin{itemize}
 \item A holistic and insightful review of ternary quantization.
  
 \item A detailed comparison of formulation and optimization during the ternary quantization process.
 
 \item Insights into the connections among different ternary quantization methods, which can be generalized to multi-bit quantization.
    
\end{itemize}

\begin{figure}
     \centering
     \begin{subfigure}[b]{0.47\textwidth}
         \centering
         \includegraphics[width=\textwidth]{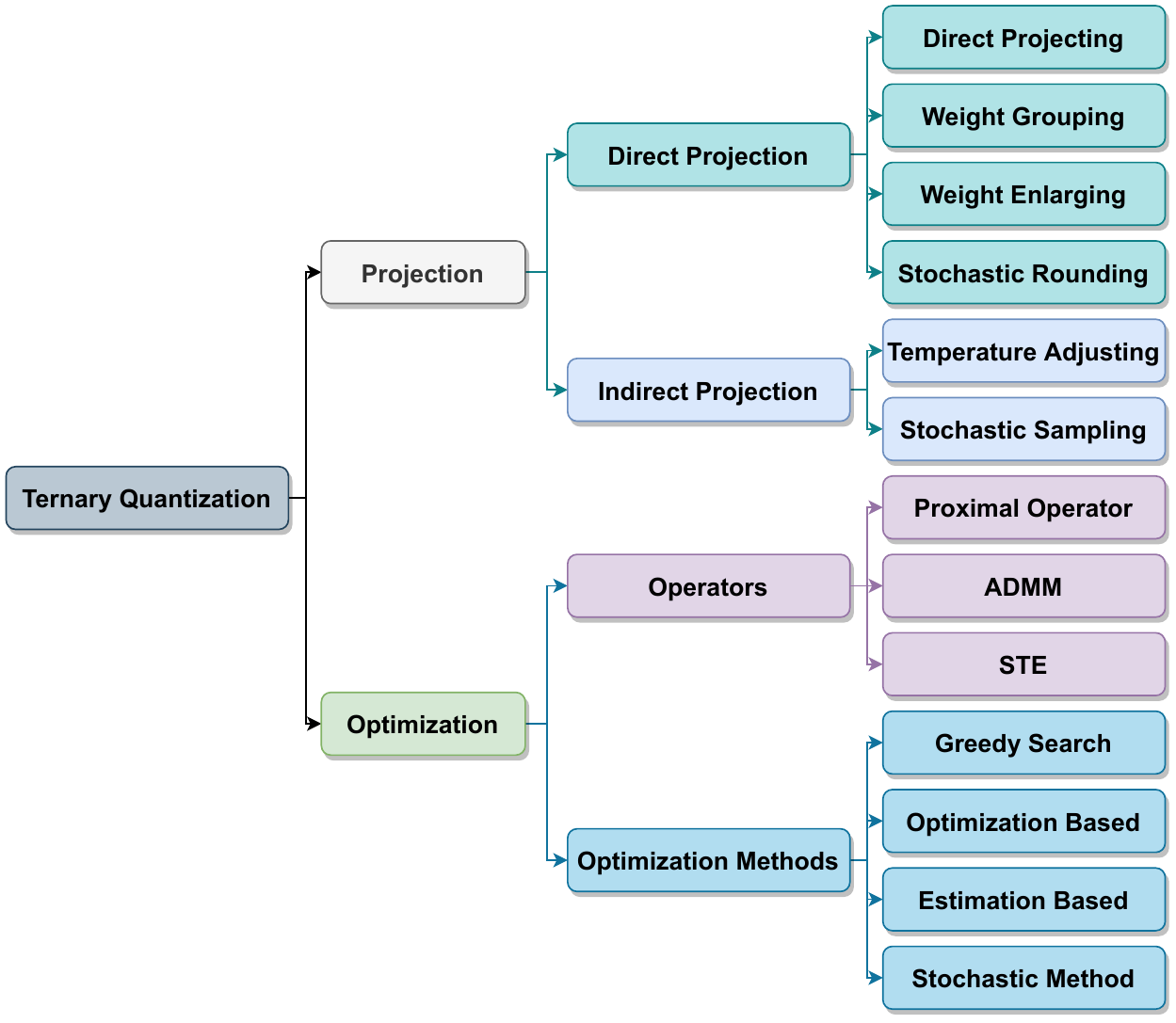}
         \label{fig:three sin x}
     \end{subfigure}
    \caption{The overall structure of this ternary quantization survey.}
    \label{img:overall}
\end{figure}

The overall structure of this survey is shown in Figure \ref{img:overall}. Section~\ref{sec:ternary_neural_nets} is a brief introduction to ternary quantization. Section~\ref{sec:proj func} describes how ternary quantization is formulated based on projection functions. Section~\ref{sec:opt method} summarizes various optimization methods from the perspective of the proximal operator. Recommendations on potential research topics on ternary quantization and conclusions are in Sections~\ref{sec:future} and \ref{sec:conclusion}, respectively.

\section{Ternary Neural Networks}
\label{sec:ternary_neural_nets}
A general representation of a neural network layer is:
\begin{equation}
\small
\label{eq:spop}
    \mathbf{y}=\phi(\mathbf{w}^\top\mathbf{x}),
\end{equation}
where $\mathbf{w} \in \mathbb{R}^{n\times{m}}$, $\mathbf{x}\in \mathbb{R}^{n}$ is the input vector to the layer, $\phi$ represents a nonlinear
activation function, and $\mathbf{y}\in \mathbb{R}^{m}$ is the output feature vector.
Ternary quantization has two components, i.e., thresholding and projection:
\begin{equation}
\small
\hat{w}=\texttt{Q}(w)=\left\{\begin{aligned} 1 &: w>\Delta \\ 0 &:\left|w\right| \leq \Delta  \\-1 &: w<-\Delta, \end{aligned}\right.
\label{eq:TQ}
\end{equation} 
\begin{equation*}
    s.t.~~ \hat{w}\in\mathbf{\hat{w}}, w\in \mathbf{w}, ~\text{and}~\Delta \in \mathbb{R},
\end{equation*}
where $\Delta$ denotes the quantization threshold. The optimization objective can be defined as:
\begin{equation}
\small
\label{eq:twn}
     \underset{{\mathbf{w},\alpha}}{\operatorname{argmin}}~ J(\mathbf{w})= \|\mathbf{w}-\alpha\mathbf{\hat{w}}\|_2,
\end{equation} 
where $\alpha$ denotes a scaling factor. Given $\Delta$, $\alpha$ has a closed-form solution \cite{lei_deng_gxnor-net_2018_r30,li_ternary_2016_r48}. More details are in Section \ref{direct_projection}.  

\subsection{Brief History}
Early ternary quantization work aims to deploy models on dedicated hardware. The earliest literature dates back to 1988, in the work of \citet{chiueh_learning_1988_r56}. Due to hardware limitations, the weights must be quantized into two or three levels in order to program weights onto VLSI chips. The author of \cite{chiueh_learning_1988_r56} studies several learning algorithms to project continuous weights into ternary values. It compares six schemes for training a three-layer network with random initialization. They do not explain specific implementation details or outline the workflow. Based on their algorithms and overview, they use a genetic algorithm combined with gradient descent and simulated annealing to solve a ternary weight optimization problem. They claim that the genetic algorithm outperforms other methods. 

In 1990, the work of \citet{marchesi_design_1990_r55} provided a detailed ternary quantization training process. One difference
from most modern works is that they do not keep the original
copy of full-precision weights but quantize the model immediately after the gradient update. Research efforts were made in the subsequent five years to extend this work. However, the network is elementary, and the data set is small \cite{curkowicz_back-propagation_1994_r54,hoskins_vlsi_1995_r53}. Later, \citet{hwang_fixed-point_2014_r52} tried to use a method similar to Straight-Through Estimator (STE) \cite{bengio2013estimating} for quantization. The network is smaller and only has three layers. 

Until 2016, with the emergence of Ternary Weight Networks (TWN) \cite{li_ternary_2016_r48}, ternary quantization began to be used for large models (VGG \cite{simonyan2014vgg}, ResNet\cite{he2016resnet}) and large datasets (ImageNet \cite{deng2009imagenet}). Therefore, those methods designed for simple data and models are no longer applicable. In the early days, TWN and Trained Ternary Quantization (TTQ) \cite{zhu_trained_2016_r41} were relatively intuitive, and the overall accuracy was different from the present ternary quantization methods. Since then, several hardware-related implementations \cite{esser_convolutional_2016_r49,adrien_prost-boucle_scalable_2017_r44,ando_brein_2017_r43,mishra_wrpn_2017_r38,lei_deng_gxnor-net_2018_r30,di_guglielmo_compressing_2020_r19,jain_tim-dnn_2020_r16,laborieux_low_2020_r14,chen_fatnn_2021_r8} have emerged on the basis of TWN and TTQ work. These methods focus on designing hardware; the network is no longer three layers.

Since 2017, some new branches of ternary quantization have been proposed. For example,  ternary constrained optimization \cite{leng_extremely_2017_r39}, knowledge distillation for low-bit models \cite{mishra_apprentice_2017_r42}, matrix factorization \cite{wang_fixed-point_2017_r35}, partitioning \cite{zhou_incremental_2017_r34,zhou_explicit_2018_r26.5,wang_two-step_2018_r27}
, stochastic quantization \cite{achterhold_variational_2018_r33.5,louizos_relaxed_2018_r29.5,shayer_learning_2018_r29}
, smooth transition \cite{wang_hitnet_2018_r28} and widening \cite{mishra_wrpn_2017_r38}. We will elaborate on the development of these methods in the upcoming sections.


\subsection{An Overview of Ternary Optimization}
The scaling factor $\alpha$ and threshold $\Delta$ in Eq.~\eqref{eq:twn} can be non-symmetric. 
To obtain ternary weights $\{-\alpha,0,\alpha\}$, most ternary quantization works focus on designing the projection function to make deployment on hardware devices possible. However, ternary quantization usually comes with three problems: the difference between $\mathbf{w}$ and $\mathbf{\hat{w}}$ is significant, the model can get stuck in local minima \cite{gholami2021survey,chiueh_learning_1988_r56}, and the projection brings zero gradients. Therefore, one of the purposes of ternary quantization is to improve the accuracy of the quantized model by narrowing the gap between $\mathbf{\hat{w}}$ and $\mathbf{w}$ \cite{li_ternary_2016_r48}. In addition, since the projection function will neutralize the gradient update once the magnitude values of updated $\mathbf{w}$ are less than the $\Delta$, the model will stall at the local minimum. Rescaling $\mathbf{w}$ with $\alpha$ can help jump out of these local minimums and thus improve the accuracy of the quantized model \cite{curkowicz_back-propagation_1994_r54}. Therefore, $\alpha$ and $\Delta$ can reduce the gap between $\mathbf{\hat{w}}$ and $\mathbf{w}$ \cite{li_ternary_2016_r48}, and jump out of local minimums.

Early work of \citet{chiueh_learning_1988_r56} (1988) proposed to jump out of the local minimum by random initialization. The disadvantage is that the old version of weights cannot be used for subsequent training. Therefore, \cite{chiueh_learning_1988_r56} further combines the genetic algorithm (GA) to generate a new weight by mixing the stalled weights.  The work of \citet{marchesi_design_1990_r55} was the first one that proposed a scaling factor to alleviate this local minimum problem. The author of \cite{marchesi_design_1990_r55} projects the full-precision weights to \{0, 1, -1\}, and then uses $\alpha$ and $\Delta$ to adjust the direction and magnitude of the quantized weights. Their model has only three layers with 132 units. Thus, they use greedy search to obtain the scaling factor $\alpha$ and threshold $\Delta$. Another noteworthy point is that the work of \cite{marchesi_design_1990_r55} uses full-precision weights only for initialization. Then the gradients are updated directly on the ternary weights, meaning that the full-precision weights are not helpful for the following optimization process. 

From the work of \cite{hwang_fixed-point_2014_r52}, ternary quantization methods tend to use dual-averaging sub-gradient algorithms \cite{nesterov2009primal_dual_averaging}, i.e., updating gradients to full-precision copies rather than ternary weights, which can alleviate the local minima problem caused by projection functions. In this way, the gradients can accumulate on full-precision weights until the magnitude values of $\mathbf{w}$ are greater than the $\Delta$. From another perspective, the intuition of the full-precision update approach is to find the optimal ternary weights near the optimal full-precision weights. Most of the works after 2014 performed gradient updates on full-precision weights and used pre-trained models as initialization when using large-scale datasets, e.g., ImageNet.

After 2014, more and more works take Eq.~\eqref{eq:twn} as a base formula for ternary quantization, which focuses on two parameters, one is the scaling factor $\alpha$, and the other is the threshold $\Delta$. Intuitively, the purpose of these two parameters is actually to make $\mathbf{w}$ and $\mathbf{\hat{w}}$ closer. For example, $\Delta$ and $\alpha$ affect the direction and magnitude of $\mathbf{w}$. The model weights are scaled by $\alpha$, then projected by $\Delta$. The scaling factor $\alpha$ can be a fixed value \cite{li_ternary_2016_r48} or an optimized target \cite{zhu_trained_2016_r41}. Usually, the threshold $\Delta$ divides $\mathbf{w}$ into zero and non-zero values. It is worth noting that $\Delta$ is usually obtained by estimation rather than optimization. Given $\Delta$, $\alpha$ has a closed-form solution (Eq.~\eqref{eq:alpha}). 
In addition, many works optimize $\alpha$ and $\Delta$ iteratively (See section \ref{sec:admm}). Given an estimated $\Delta$, the scaling factor is first calculated by Eq.~\eqref{eq:alpha} (like the method in the 90s), and the $\Delta$ is estimated again based on the newer $\alpha$. This process can be repeated several times until the distance between $\mathbf{w}$ and $\mathbf{\hat{w}}$ is small enough. Therefore, we conclude that the scaling factor $\alpha$ and threshold $\Delta$ can optimize the distance between $w$ and $\hat{w}$ and jump out of local minima.

Although $\alpha$ and $\Delta$ can help with ternary projection, projecting continuous variables into discrete spaces leads to the optimization problem of non-differentiable and zero gradients. From the perspective of proximal optimization \cite{parikh2014proximal}, proximal gradient descent can be used to approximately solve these problems. The widely used STE also can be seen as a special case of proximal optimization \cite{parikh2014proximal}. Most of the optimization process keeps a full-precision copy of the model weights and performs optimization based on the gradients produced by quantization.

This survey covers the ternary quantization works after 2014, as the previous ones are based on small-size networks. The structure of this survey is organized according to the projection function and optimization of ternary quantization.

\section{Projection Function}
\label{sec:proj func}
The purpose of the projection function is to produce discrete weight values. Unlike other works that categorize the projection functions into deterministic and stochastic ones, we classify them according to the state of the weights in the forward pass. The following content in this section includes a summary of direct and indirect projection functions. (Note that from this section, the equations in this survey from other works will follow the definitions of the original work, and we will provide the accurate source of such equations.)
\subsection{Overview of The Projection Function} %
For most projection-based ternary quantization methods, the backpropagation gradients are estimated values. Although STE \cite{bengio2013estimating} or Hinton's slides are the most popular, similar methods were used in as early as 1994, in the work of \cite{marchesi_design_1990_r55}. Different from the STE, in the early works, the weights are quantized by the projection immediately after gradient updates. Obviously, the gradient update will be neutralized by projection functions, which brings the problem of getting stuck in local minima \cite{chiueh_learning_1988_r56} during training.
Until recently, some works have shown that the dual-averaging method \cite{nesterov2009primal_dual_averaging} can alleviate this problem of weight neutralization. The dual-averaging method in the ternary quantization scenario can be interpreted as using projected (quantized) weight values to produce gradients for updating full-precision weights during backpropagation, which is now a widely applied quantization standard.

The purpose of indirect projection methods is to relieve inaccurate gradients by employing a gradual projection process. Some works perform single-element-wise projection, for example, mixing ternary and full-precision values to create hybrid values and gradually adjust the portion of ternary values \cite{bai_proxquant_2018_r33} or using a temperature-based function to adjust the discrete degree \cite{yang_quantization_2019_r22,louizos_relaxed_2018_r29.5}. Other works gradually mix ternary, and full-precision values based on a score function \cite{zhou_incremental_2017_r34} or randomly \cite{fan2020noisequant}. In addition to the gradual projection methods, stochastic methods either sample discrete weights over the multinomial distribution or add normal-distributed noise to discrete weights to create continuous ones \cite{shayer_learning_2018_r29}. However, the sampling efficiency of those methods could be a challenge for deep neural networks.

\subsection{Direct Projection}


Discrete projection uses projection functions to create non-differentiable discrete values in the forward pass. It can be divided into fully and partially discrete methods. 
Partially discrete methods include group-wise and random discretization. What needs to be considered is how the gradient can be updated to the weight during backpropagation and the solution of corresponding $\alpha$ and $\Delta$. 
The key factors of ternary quantization are to find the proper scaling factor $\alpha$, threshold $\Delta$, and sub-gradient to minimize the difference between $\mathbf{\hat{w}}$ and $\mathbf{w}$. 
\subsubsection{Direct Projecting}
The earliest direct projection method was mentioned in 1988 \cite{chiueh_learning_1988_r56}, without any details on the training process. 

The first detailed ternary training appeared in 1990 (Eq. (5-8) of \cite{marchesi_design_1990_r55}):
\begin{equation}
\small
\mathrm{E}_{\mathrm{h}}(\mathrm{A})=\sum_{\mathrm{s}} \sum_{\mathrm{k}} 
\sum_{\mathrm{j}=1}^{\mathrm{N}_{\mathrm{s}-1}}
\left(\mathbf{w'}_{\mathrm{kj}}^{(s)}
-\mathrm{A}^{-1} <\mathbf{w}_{\mathrm{kj}}^{(s)} \mathrm{A}>\right)^{2}
\end{equation}
\begin{equation}
\small
\mathbf{w''}_\mathrm{kj}^{(s)}=\left\langle A_{h} \mathbf{w'}_\mathrm{kj}^{(s)}\right\rangle.
\end{equation}
\begin{equation}
\small
\label{eq:ref55}
\mathbf{w''}_\mathrm{kj}^{(s)}(t+1)=\left\langle \mathbf{w''}_\mathrm{kj}^{(s)}(t)+\eta \Delta_{k}^{(s)}\right\rangle
\end{equation}
where $A$ is the scaling factor, $\langle \cdot \rangle$ denotes rounding, and $\Delta$ is the gradient. We can see that the work of \cite{marchesi_design_1990_r55} tries to find a proper scaling factor $A$ that minimizes the distance between the scaled  $\mathbf{\hat{w}}$ and $\mathbf{w}$. Then use the scaled $\mathbf{\hat{w}}$ for the forward pass. The difference between their method and STE is that Eq.~\eqref{eq:ref55} rounds $\mathbf{w}$ directly after the gradient update. 

\citet{hwang_fixed-point_2014_r52} project $\mathbf{w}$ into discrete space in the forward pass to produce gradients and keep a full-precision copy of weight values for gradient updates. It also extends the work to higher bit quantization scenarios (Fig 3 and Eq.(5-8) in \cite{hwang_fixed-point_2014_r52}):
\begin{equation}
\small
\begin{array}{c} {w}_{i j, n e w}=w_{i j}-\alpha\left\langle
\frac{\partial E}{\partial w_{i j}}\right\rangle
\\ w_{i j, n e w}^{(q)}=Q_{i j}
\left(w_{i j, n e w}\right)
\\ Q(x)=\Delta \operatorname{sgn}(x) \cdot \min \left\{ \left\lfloor \frac{|x|}{\Delta}+0.5\right\rfloor, \frac{M-1}{2}\right\}\end{array}
\end{equation} 
where $\frac{M-1}{2}$ is the number of bits, $\Delta$ is optimized in a greedy manner, and $\left\langle\cdot\right\rangle$ returns averaged values. The gradient $\frac{\partial E}{\partial w_{i j}}$ is used to update the full-precision $\mathbf{w}$, which is actually the use of STE. \citet{sung_resiliency_2016_r45} use the same quantization method to study the performance gap between the floating point and the ternary weights from the perspective of the model complexity. 

\citet{li_ternary_2016_r48}, the authors of TWN, use STE for quantization in the subsequent work. In their work, the threshold $\Delta$ is an estimated value, and the scaling factor $\alpha$ has a closed-form solution (Eq.~\eqref{eq:alpha}). \citet{diwen_wan_tbn_2018_r32} apply TWN to activation quantization with binary weight values. \citet{penghang_yin_training_2016_r46} introduce two scaling factors with an estimated $\Delta$ to obtain more accurate ternary values. \citet{penghang_yin_quantization_2017_r36} restrict the scaling factor $\alpha$ with the form of power-of-two and provide a lower bound of $\alpha$; and it also estimates the $\Delta$ like other works \cite{zhu_trained_2016_r41,li_ternary_2016_r48} do. The work of \cite{alemdar_ternary_2017_r50,adrien_prost-boucle_scalable_2017_r44} uses a greedy Dichotomic search to find the optimized $\Delta$, and defines a score function to measure the difference between $\mathbf{w}$ and $\hat{\mathbf{w}}$ (Eq.~(3) in \cite{alemdar_ternary_2017_r50}). \citet{jie_ding_three-means_2017_r40} propose a preprocessing method based on  k-means to assign each weight value to cluster centers, i.e., {$\pm \alpha, 0$}. The scaling factors $\pm\alpha$ play the role of the threshold $\Delta$. The author uses a method similar to TWN to obtain $\alpha$. The final quantization is fulfilled by regular STE. 


\subsubsection{Weight Grouping}

Grouping methods gradually increase the portion of the projected full-precision weights, i.e. $\texttt{Q}(\mathbf{w})$. The advantage of grouping is that using mixed precision weights can bring smaller errors than fully-quantized weights and thus produce more accurate gradients. The error can be measured by: 
\begin{equation}
\small
\label{eq:grp}
\begin{array}{c}
    \|\mathbf{y}-\mathbf{x}^\top \mathbf{w}\|_2^2 \le \|\mathbf{y}-\mathbf{x}^\top (r \mathbf{w}+ (1-r)\texttt{Q}(\mathbf{w}))\|_2^2 \\
    \le \|\mathbf{y}-\mathbf{x}^\top \texttt{Q}(\mathbf{w})\|_2^2, 
\end{array}
\end{equation}
where $r$ controls the projected portion of $\mathbf{w}$. One of the extreme cases can be found in \cite{lei_memory_2020_r13}:
\begin{equation*}
\begin{array}{c} \min _{w_{j}} L= \frac{1}{n} \sum_{i=1}^{n} \max \left(0,1-y_{i} \alpha\left(\mathbf{w}_{\neg j}^{\top} \mathbf{z}_{i}+w_{j} z_{i j}\right)\right) \\ +\lambda \alpha^{2} w_{j}^{2}+\lambda \alpha^{2} \mathbf{w}_{\neg j}^{\top} \mathbf{w}_{\neg j}, \end{array}
\end{equation*}
\begin{equation}
\small
    \label{eq:ref13}
    \text { s.t. } \quad w_{j} \in\{-1,0,1\}
\end{equation}
where $\mathbf{z}$ is the input, $L$ denotes the hinge loss, and $\mathbf{w}_{\neg j}$ denotes the full-precision weight vector with $j$-th entry equaling zero. Eq.~\eqref{eq:ref13} indicates that:
one can perform ternary quantization element by element to obtain optimal ternary weights. Or one can do that reverse; for example, many other methods \cite{zhou_incremental_2017_r34,hu_cluster_2019_r24,wang_two-step_2018_r27} fix ternary weights but update full-precision weights. The fixed portion of ternary weights is gradually increased during training. The work of \citet{zhou_incremental_2017_r34} uses weight magnitude as a reference to fix the ternary weights. Whereas \citet{hu_cluster_2019_r24} use column-wise optimization, i.e., optimizing one column by fixing other columns. \citet{wang_two-step_2018_r27} and \citet{yang_quantization_2019_r22} perform element-wise and layer-wise quantization, respectively. 

In addition, \citet{esser_convolutional_2016_r49} introduce a full-precision buffer zone around $|\pm0.5|$ and use $h$ to control the width of that zone:
\begin{equation}
\small
w(t)=\left\{\begin{array}{ll}-1 & \text { if } w_{h}(t) \leq-0.5-h, \\ 0 & \text { if } w_{h}(t) \geq-0.5+h \vee w_{h}(t) \leq 0.5-h, \\ 1 & \text { if } w_{h}(t) \geq 0.5+h, \\ w(t-1) & \text { otherwise }\end{array}\right.
\end{equation}
By setting a continuous zone $[\pm0.5,\pm h]$, once the weight values in the zone obtain enough gradients, they will have a chance to jump out and become discrete. However, this design cannot guarantee that all weights can jump out.
\citet{naveen_mellempudi_ternary_2017_r37} combine the concept of grouping with the quantization scheme of TWN. More specifically, the $\mathbf{w}$ is divided into $N$ orthogonal subsets and the optimal $\alpha$ and $\Delta$ are optimized through brute force search (minimizing the $L2$ distance between $\mathbf{w}$ and $\mathbf{\hat{w}}$). The authors of \citet{naveen_mellempudi_ternary_2017_r37} also find that grouping through input channels can obtain better ternary quantization results. SYQ (\cite{faraone_syq_2018_r31.5}) proposes a similar method as \cite{naveen_mellempudi_ternary_2017_r37} but with a different grouping scheme and optimizes group-wise $\alpha$ with SGD \cite{zhu_trained_2016_r41}. \citet{wang_fixed-point_2017_r35} decompose the full-precision weight matrix into three subsets in a greedy manner and performs ternary quantization on those subsets; however, the training details are not clear. INQ (\cite{zhou_incremental_2017_r34}) quantizes and freezes the full-precision weights with a larger magnitude. The rest flexible weight values are further optimized and gradually quantized. The work of \citet{zhou_explicit_2018_r26.5} defines ternary centers and then gradually increases the quantized portion around the ternary centers. \citet{cavigelli_rpr_2020_r21} randomly divide the weights into triples and full-precision groups. First, the ternary part is frozen, and the full-precision part is updated. Then the full-precision is frozen, and the ternary is updated to full-precision. Repeat this process and simultaneously increase the proportion of $\mathbf{\hat{w}}$ until quantization is complete.


\subsubsection{Weight Enlarging}
Duplicating the layer weights \cite{he_optimize_2019_r25} and increasing layer width \cite{mishra_wrpn_2017_r38} can increase the capacity of the model. We name this kind of operation enlarging.

\citet{mishra_wrpn_2017_r38} study the impact of increasing model redundancy, e.g., increasing the layer width and the number of ternary filters, on the accuracy of ternary quantization. \citet{he_optimize_2019_r25} study adding additional ternary residual layers to enhance the model performance. Each layer has multiple copies of the weights, and those copies are empirically assigned with different $\Delta$. \citet{xu_soft_2020_r11} produce the ternary by adding two binary weights with scaling factor $\alpha$, which avoids the uncertainty of $\Delta$: $\{\alpha,-\alpha\}+\{\alpha,-\alpha\}=\{2\alpha,0,-2\alpha\}$.
Similarly, \citet{li_trq_2021_r5} stack two ternary weights with coefficients to produce new ternary values.

\citet{dbouk_dbq_2020_r20} propose a separable ternary structure. Two ternary weights are stacked together as mixed weights during training. The mixed weights can be split into two ternary weights for fast inference. During inference, the two inner products of a layer can be calculated separately but at the cost of doubled computation overhead. \citet{kim_binaryduo_2020_r15} decouple ternary activations into two binary activations by adjusting the bias of the BatchNorm layer. The ternary activations are decomposed into the sum of the two binary ones. 

To summarize, enlarging is proposed to compensate for the reduced model capacity caused by ternary quantization. With the use of enlarging, the performance can be improved, but additional computational overhead will also be needed.

\subsubsection{Stochastic Rounding}
\label{StochasticRounding}

The above-mentioned methods sometimes are usually categorized as deterministic methods. For stochastic methods, an earlier article on stochastic ternary quantization is proposed by \citet{lin_neural_2016_r47}, which is an extended work of binary quantization \cite{courbariaux2015binaryconnect}. The works of \citet{laborieux_low_2020_r14} and \citet{kim_binaryduo_2020_r15} show that randomized or stochastic rounding can provide unbiased discrete projection. \citet{lin_neural_2016_r47} apply stochastic ternary quantization in the forward pass, and the power-of-two quantization is applied to the input to eliminate multiplications during backpropagation. \citet{alemdar_ternary_2017_r50} use a pre-trained full-precision (teacher) model to guide the same structure ternary (student) model. It randomly quantizes the teacher activations and then enables the student to imitate these activations by using layer-wise algorithms. The activations are quantized before the weights. 

Unlike commonly used methods that obtain gradients in discrete space and update weight in continuous space, in the work of \citet{lei_deng_gxnor-net_2018_r30}, the gradients are stochastically projected into discrete space to update ternary weights. There are six possible transitions for the weight values in total, i.e., $\{-1\} \rightarrow \{0,1\}$, $\{0\} \rightarrow \{-1,+1\}$, and $\{1\}  \rightarrow \{-1,0\}$ (see Fig. 3 in \cite{lei_deng_gxnor-net_2018_r30}). The probability of projection depends on the magnitude and direction of the gradients.

In summary, stochastic rounding selects one of the two closest grid points with probability depending on different measurements, e.g. weight magnitude. Although stochastic rounding can reduce the biased gradient in projection, the sampling speed is slow [ref]. Therefore the efficiency of generating weights is very low, especially when the number of weight elements is large, and the training and inference efficiency of the model will be reduced. 

\subsection{Indirect Projection}
The above-mentioned direct projection methods suffer from non-differentiability. The indirect projection methods try to partially or gradually bypass such non-differentiability. In other words, they try to train a quantized model in the continuous space as much as possible. The indirect projection methods can be categorized into gradual and stochastic discrete projection methods.

\subsubsection{Temperature Adjusting}
To gradually adjust the discreteness, a common approach is introducing a temperature parameter to a projection function, e.g., \textcolor{black}{tanh \cite{gong_differentiable_2019_r26}, Sigmoid \cite{yang_quantization_2019_r22}, and GumbelSoftmax \cite{louizos_relaxed_2018_r29.5}}. The higher the temperature, the closer to the step function. The temperature parameter controls the steepness. The temperature adjustment can adapt the strength of discreteness during the quantization.

In the work of \citet{yang_quantization_2019_r22} and \citet{dbouk_dbq_2020_r20}, the continuous weights pass through a Sigmoid function and become discrete when gradually increasing the temperature of the Sigmoid function:
\begin{equation}
\small
 \sigma(T x)=\frac{1}{1+\exp (-T x)}.
\end{equation}
Increasing the temperature $T$ allows the weight values to be gradually converted from continuous to discrete. \citet{wang_hitnet_2018_r28} extend the work of TTQ to RNN and uses a sloping factor to adjust the steepness of activation for quantization. Unlike TWN uses an indicator function directly, the work of \citet{bai_proxquant_2018_r33} proposes a soft threshold function to fulfill the quantization. More specifically, they propose $L1$ and $L2$ regularizers to adjust the degree of quantization. Then, a gradually increased $\lambda$ is used to adjust the strength of regularizers to convert the full-precision weights to discrete ones.  Therefore the ProxQuant \cite{bai_proxquant_2018_r33} can be seen as a gradual quantization method. The GumbelSoftmax-based method, like \cite{louizos_relaxed_2018_r29.5}, also involves a gradually changed temperature parameter to convert the continuous weights into discrete ones.
\subsubsection{Stochastic Sampling}
There are two types of stochastic quantization methods. One stochastically projects continuous weight into discrete (Section \ref{StochasticRounding}). And the other stochastic quantization method makes discrete weights continuous again by adding noise to facilitate optimization \cite{achterhold_variational_2018_r33.5}, or gradually converting stochastic projection from continuous space into discrete space \cite{louizos_relaxed_2018_r29.5}.
In the work of \citet{shayer_learning_2018_r29}, the multinomial distribution is implemented with two Sigmoid functions, and the layer weights are used as the probability parameter, however, at the cost of doubling the weight size. The mean and variance of $\mathbf{\hat{w}}$ are calculated after sampling over multinomial distribution. With the mean and variance value, the noise $\mathcal{N}(0, I)$ is used to generate continuous weights in the forward pass.
The discrete weights are obtained during inference by directly sampling over the multinomial distribution (For more details, please see Section \ref{indirect_projection}). \citet{louizos_relaxed_2018_r29.5} use Gumbel-Softmax and temperature parameters. It can be seen as an enhanced version of the LR-net \cite{shayer_learning_2018_r29}, as it involves a gradually changing temperature as in the Sigmoid-based method. The final form of the weight is discrete, whereas the LR-net directly samples discrete weight during inference, which is inefficient.
Stochastic sampling can sample multiple weights to form an ensemble model, however, the overall accuracy could be better.

In summary, the indirect projection has the issue of low running efficiency. The temperature adjusting methods use Sigmoid functions to gradually convert continuous weight to discrete. Accuracy is high, while the training is time-consuming. 
For example, the quantization method of \cite{yang_quantization_2019_r22} performs time-consuming layer-wise quantization and has three phases: weight quantization training, activation quantization training with quantized weights, and fine-tuning with both quantized activations and weights together. The training time depends on the depth of the network. 
The stochastic methods suffer from inefficient sampling -- the more parameters the network has, the slower the sampling speed will be.

\section{Optimization Methods}
\label{sec:opt method}
The optimization of ternary quantization can be categorized in many ways, for example, optimization-based, estimation-based, and greedy-search-based. One can even use alternative training schemes, knowledge distillation, or additional loss terms to help with the optimization process. 
We summarize the connections among various forms and use proximal mapping to unify the direct projection optimization methods. 

\subsubsection{Proximal Operator}
Since proximal operators can be viewed as generalized projections \cite{parikh2014proximal}, we use the proximal operator to unify ternary quantization methods. In this survey, the proximal operator under ternary settings is:
\begin{equation}
\small
\operatorname{prox}_{I_{\texttt{Q}}}(w)=\underset{x\in\mathbb{R},\alpha\in\mathbb{R}}{\operatorname{argmin}}\left(I_{\texttt{Q}}(x)+\frac{1}{2\lambda}\|w-\alpha x\|_{2}^{2}\right),
\end{equation}
where $\alpha$ is a scaling factor. If $I_{\texttt{Q}}(x)$ is an indicator function:
\begin{equation}
\small
I_{\texttt{Q}}(x)=\left\{\begin{array}{ll}0 & x \in \texttt{Q}(x), \\ +\infty & x \notin \texttt{Q}(x),\end{array}\right.
\end{equation} 
then the proximal operator of $I_{\texttt{Q}}$ reduces to Euclidean projection $P(\cdot)$ onto a closed nonempty convex set $\texttt{Q}(x)$:
\begin{equation}
\small
\label{eq:proj_argmin}
P_{\texttt{Q}}(w)=\underset{\alpha, x \in \texttt{Q}(x)}{\operatorname{argmin}}\|w-\alpha x\|_{2}.
\end{equation}
Eq.~\eqref{eq:proj_argmin} is equivalent to Eq.~\eqref{eq:twn}.
The optimization process of ternary quantization can be seen as a Euclidean Projection, i.e., projecting full-precision weights into discrete sets $\texttt{Q}(\cdot)$.

\subsubsection{Alternating Direction Method of Multipliers}
\label{sec:admm}
Alternating minimization is commonly used when optimizing with two variables. Ternary quantization can be seen as constrained convex optimization, i.e., a special case of the Alternating Direction Method of Multipliers (ADMM) \cite{boyd2011ADMM}. We take the ADMM as a general example to summarize the widely used alternating training scheme in ternary quantization.
Consider a generic constrained convex optimization: 
\begin{equation}
\small
\begin{array}{ll}\operatorname{minimize} & f(w) \\ \text { s.t. } & w \in \texttt{Q}(w).\end{array}
\end{equation}
The ADMM form of the above is:
\begin{equation}
\small
\begin{array}{ll}\text { minimize } & f(w)+I_{\texttt{Q}}(z) \\ \text { s.t. } & w-z=0.\end{array}
\end{equation}
The training algorithm is described below:
\begin{equation}
\small
\begin{aligned}w^{k+1} &:=\operatorname{prox}_{f}(w)=\underset{w}{\operatorname{argmin}}\left(f(w)+\frac{1}{2\lambda}\left\|w-z^{k}+u^{k}\right\|_{2}^{2}\right),\\z^{k+1} &:=\operatorname{prox}_{I_{\texttt{Q}}}(w)=P_{\texttt{Q}}\left(w^{k+1}+u^{k}\right), \\
u^{k+1} &:=u^{k}+w^{k+1}-z^{k+1}, \end{aligned}
\end{equation}
where $k$ denotes the training iteration.
ADMM is most useful when the proximal operators of $f$ and ${I_{\texttt{Q}}}$ can be efficiently evaluated, but the proximal operator for $f+{I_{\texttt{Q}}}$ is not easy to evaluate (See: page 154 of Proximal Algorithms\cite{parikh2014proximal}). Therefore, one can easily use the alternating minimization method to perform quantization \cite{hwang_fixed-point_2014_r52,wang_two-step_2018_r27,lei_memory_2020_r13,yang_quantization_2019_r22,hu_cluster_2019_r24,hou_loss-aware_2018_r31,bai_proxquant_2018_r33,wang_fixed-point_2017_r35}. For example, given a specific $\Delta$ and $\alpha$ to optimize $\operatorname{prox}_{f}(w)$, then optimize $\operatorname{prox}_{I_{\texttt{Q}}}$ to get a newer version of $\Delta$ and $\alpha$, repeat this process until the model convergence.

\subsubsection{Straight-Through Estimator}
The optimization process of ternary quantization with a Straight-Through Estimator is equivalent to a ternary projection function $P_{\texttt{Q}}(w)$ with dual-averaging \cite{nesterov2009primal_dual_averaging}:
\begin{equation}
\small
\left\{\begin{array}{l}{{\hat{w}}}_{t}:=P_{\texttt{Q}}(w)=\underset{\alpha, x \in \texttt{Q}(x)}{\operatorname{argmin}}\|w-\alpha x\|_{2}, \\ w_{t+1}=w_{t}-\eta_{t} {\nabla} L\left({\hat{w}}_{t},\right)\end{array}\right.
\end{equation} where $t$ denotes the training iteration, and $\eta_{t}$ denotes the learning rate. When a=1, Eq.~\eqref{eq:proj_argmin} becomes very common quantization function, such as \cite{laborieux_low_2020_r14,di_guglielmo_compressing_2020_r19,cavigelli_rpr_2020_r21,lei_deng_gxnor-net_2018_r30,ando_brein_2017_r43,lin_neural_2016_r47,esser_convolutional_2016_r49,alemdar_ternary_2017_r50,hwang_fixed-point_2014_r52}. Many hardware-quantization articles \cite{chen_fatnn_2021_r8,laborieux_low_2020_r14,jain_tim-dnn_2020_r16,di_guglielmo_compressing_2020_r19} also prefer to use this method. However, projecting to $\{\pm1, 0\}$ widens the distance between $\mathbf{w}$ and $\mathbf{\hat{w}}$.

Considering the case of $\alpha>0$, the well-known paper TWN \cite{li_ternary_2016_r48} proposes the use of Eq.~\eqref{eq:proj_argmin}. 
In addition to the above-mentioned projection $P_\texttt{w}$, the work of TWN also uses an approximated $\Delta$ to minimize the optimization objective and then projects $\mathbf{w}$ onto $\alpha\mathbf{\hat{w}}$. Unlike the work of TWN, TTQ \cite{zhu_trained_2016_r41} takes $\alpha$ as an optimized parameter.


\subsection{Direct Projection}
\label{direct_projection}
Direct projection methods usually include $\alpha$ and $\Delta$. In the following content, the works are summarized in three aspects of how to get $\alpha$ and $\Delta$: estimation-based, optimization-based, and greedy-search-based methods.
\subsubsection{Estimation Based Methods}
The estimation-based methods usually take Eq.~\eqref{eq:proj_argmin} as the foundation. As indicated in Eq.~\eqref{eq:proj_argmin}, the purpose of estimating $\alpha$ and $\Delta$ is finding the shortest $L2$ distance between $\mathbf{w}$ and $\mathbf{\hat{w}}$. Given a fixed $\Delta$, a closed-form solution of $\alpha$ is:
\begin{equation}
\small
    \alpha=\frac{1}{\|\mathbf{w}\|_0}\sum_{i=0}^{n} |w_i|.
    \label{eq:alpha}
\end{equation}
And the matrix version of Eq.~\eqref{eq:alpha} can be found in the work of \citet{leng_extremely_2017_r39}. One of the drawbacks of estimation-based methods is that there is no explicit solution for the $\Delta$, which causes the $\alpha$ to be inaccurate. Therefore, TWN directly estimates the $\Delta$ with $\frac{0.7}{n} \sum_{i=1}^{n}|w_i|$, and TTQ uses $\Delta=0.05 \times \operatorname{max}(|\mathbf{w}|)$. As mentioned above, in Eq.~\eqref{eq:proj_argmin}, when the $\alpha$ is estimated as $1$, the quantization operation is equivalent to projecting the values to $\{0,1,-1\}$ \cite{hwang_fixed-point_2014_r52,razani_adaptive_2021_r4}. 

Although many works do not mention the form of Eq.~\eqref{eq:proj_argmin}, according to the definition of proximal mapping, any method with a projection function can be categorized into this estimation-based class. For example, in the ELBNN \cite{leng_extremely_2017_r39}, the author extends the proximal operator with augmented Lagrange, i.e., a complex projection with ADMM. ELBNN defines the matrix form of $\alpha$:
\begin{equation}
\small
\alpha=\frac{\mathbf{w}^\top\mathbf{\hat{w}}}{\mathbf{\hat{w}}^\top \mathbf{\hat{w}}}=\frac{1}{\|\mathbf{\hat{w}}\|_0}\sum_{i=0}^{n} |w_i|,
\label{eq:alpha_m}
\end{equation}
which is identical to Eq.~\eqref{eq:alpha}.

In addition the projection function, ProxQuant \cite{bai_proxquant_2018_r33} lets $\alpha=1$ in Eq.~\eqref{eq:proj_argmin}, like TWN does. The author of Proxquant propose two types of regularization terms, and the quantization function is defined as:
\begin{equation}
\small
\label{eq:proxquant}
\mathbf{\hat{w}}=\left\{\begin{array}{l}\textsc{SoftThreshold}(\mathbf{w},\mathbf{\hat{w}},\lambda), (L1~term), \\ \frac{\lambda\mathbf{\hat{w}}+\mathbf{w}}{1+\lambda}, (L2~term),
\end{array}\right.
\end{equation}
where $\lambda$ controls the portion of the quantized weights $\mathbf{\hat{w}}$. \citet{wang_fixed-point_2017_r35} mention the scaling factor $\alpha$ should be proportional to the square root of the sum number of rows and columns, however, without any details about the proposed estimation. \citet{zhou_incremental_2017_r34} use $2^n$ as the scaling factor and perform group-wise quantization in a gradual manner. \citet{diwen_wan_tbn_2018_r32} use $\Delta=\frac{0.4}{n} \sum_{i=1}^{n}|w_i|$ to produce ternary activation. \citet{zhou_explicit_2018_r26.5} add $L1$ and $L2$ regularizers to encourage the weights to move closer to ternary values. The $\alpha$ is estimated by $\texttt{mean}(\mathbf{w})+0.05\texttt{max}(\mathbf{w})$. The training process is sophisticated: the converged full-precision weights are first used to get $\alpha$, then a group-wise quantization is applied. Next, a newer $\alpha$ is obtained, and more weights are quantized group by group.
\citet{he_optimize_2019_r25} use $\alpha=\texttt{mean}(\mathbf{w})$ (TWN) and $\Delta=0.05 \times \operatorname{max}(|\mathbf{w}|)$ (TTQ) as quantization settings. In addition, they introduce an extra residual layer with independent $\alpha$. \citet{xu_soft_2020_r11} add two binary weights to obtain ternary ones and give a closed-form solution of $\alpha$. One of the advantages of the work of \cite{xu_soft_2020_r11} is that the $\Delta$ is no longer needed. The author of \cite{xu_soft_2020_r11} also proposes refined gradients (Eq.~(13) in \cite{xu_soft_2020_r11}); however, knowledge distillation is applied in their published code. Therefore, it is hard to know the result without using knowledge distillation. 

For the estimation-based works, $\Delta$ controls the number of zeros in $\mathbf{w}$. Given a $\Delta$, $\alpha$ can have a closed-form solution. However, it is hard to explicitly determine $\Delta$ due to the use of learning rate and weight-decay continuously producing a large amount of near-zero values during training. Therefore, if the estimated $\Delta$ is inaccurate, $\alpha$ is inaccurate neither. A proper sparsity can be one consideration during optimizing $\alpha$ and $\Delta$.

\subsubsection{Optimization Based Methods}
To overcome the abovementioned inaccurate $\alpha$ and $\Delta$ issue, TTQ \cite{zhu_trained_2016_r41} proposes two scaling factors that are optimized by:
\begin{equation}
\small
\frac{\partial J}{\partial \alpha^\pm}=\sum_{i=1}^{n^\pm} \frac{\partial J}{\partial w_{i}^{\pm}},
\end{equation}
where ``$\pm$'' denotes the positive or negative parts. For example, $\alpha^-$ denotes the scaling factor for negative weight values $w^-$ and is updated by $\sum_{i=1}^{n^-} \frac{\partial J}{\partial w_{i}^{-}}$, and $n^-$ denotes the number of negative values in $\mathbf{w}$. For the $\Delta$ in TTQ, it has a similar form to TWN, i.e., $\Delta=0.05\times\operatorname{max}(\mathbf{\|w\|_1})$, which reveals that $\mathbf{\|w\|_1}$ is related to layer sparsity. Therefore, the author of TTQ studies the impact of model sparsity on ternary quantization and concludes that: ``\textit{Increasing sparsity beyond 50\% reduces the model capacity too far, increasing error. The minimum error occurs with sparsity between 30\% and 50\%.}''\cite{zhu_trained_2016_r41}. \citet{mishra_wrpn_2017_r38} use the same method as TTQ, but with a widening of the network (WPRN). \citet{chen_fatnn_2021_r8} introduce a learnable $\Delta$ by replacing it with $\frac{\alpha}{2}$. \citet{faraone_syq_2018_r31.5} group the weight values and assign scaling factors to each group. Then those fine-grained scaling factors are optimized by SGD as TTQ does. 
\citet{yang_quantization_2019_r22} formulate the quantization with a Sigmoid function and a temperature coefficient $T$:
\begin{equation}
\small
    \texttt{Q}(w)=\frac{1}{1+exp(T\times (\beta w + b))},
\end{equation}
 where $\beta$ and $b$ are learnable coefficients for the quantization parameters $w$. In this way, each parameter can be optimized by SGD (Eq~(8-11) of \cite{yang_quantization_2019_r22}). 
\citet{li_rtn_2019_r23} adopt the same learnable coefficients as \cite{yang_quantization_2019_r22} by applying:
$
\bar{w}=\beta w+b,
$
where $\beta$ and $b$ are both learned via STE, and then performing ternary quantization. However, \cite{li_rtn_2019_r23} does not provide detailed gradient update rules as \cite{yang_quantization_2019_r22} does. \citet{li_trq_2021_r5} stack the residual values between the ternary $\mathbf{\hat{w}}$ and $\mathbf{w}$ to obtain a ternary projection function which has the output of $\{-2\alpha, 0, 2\alpha\}$ and does not need $\Delta$ for thresholding. Moreover, it refines the gradient of $\alpha$ and can be easily extended to arbitrary bits (see Eq.~(13-16 and 19) of \cite{li_trq_2021_r5}). 

\citet{razani_adaptive_2021_r4} extend the work of TTQ by adding loss regularization terms to mix binary and ternary quantization:
\begin{equation}
\small
\min \left(|| w|+\alpha|^{p},|| w|-\alpha|^{p}, \tan (\beta)|w|^{p}\right)
\end{equation}
By adjusting $\beta$ from $\frac{\pi}{4}$ to $\frac{\pi}{2}$, they can control the strength of the ternary quantization (Fig. 2 in \cite{razani_adaptive_2021_r4}). However, $p$ is empirically set to $1$, and no details are provided regarding optimizing $\beta$ and $\alpha$. \citet{hu_cluster_2019_r24} push the full-precision weights closer to the cluster center (ternary values) by adding regularization terms in the objective function and then using STE for ternary quantization. \citet{hou_loss-aware_2018_r31} add the second-order information (Hessian matrix) term to the objective function to adjust the gradient updates.

\subsubsection{Greedy Search Based Methods}
Greedy search method is another way to obtain $\alpha$. The works of \cite{marchesi_design_1990_r55,curkowicz_back-propagation_1994_r54,alemdar_ternary_2017_r50} use greedy search methods to determine the scaling factor $\alpha$ within a specified range, whereas \citet{hwang_fixed-point_2014_r52} directly search $\alpha$ without a range. \citet{alemdar_ternary_2017_r50} and \citet{adrien_prost-boucle_scalable_2017_r44} use a half-interval search algorithm to find the best $\pm\alpha$. \citet{jie_ding_three-means_2017_r40} uses k-means to continuously update the ternary center of $\mathbf{w}$ after each gradient update.  The values of $\mathbf{w}$ are updated with their corresponding center. Then newer centers of $\mathbf{w}$ are calculated for the next training iteration. To make the estimation of $\alpha$ and $\Delta$ in TWN more accurate, \citet{naveen_mellempudi_ternary_2017_r37} use two sets of $\alpha$ and $\Delta$ to fulfill quantization and apply brute-force search to obtain them. Specifically, each input channel has an independent scaling factor $\alpha$. \citet{penghang_yin_quantization_2017_r36} use the power-of-two scaling factor to improve performance further. When the quantization bit is greater than $2$, the $\Delta$ adopts a fixed form of $0.75\times max (||w||_1)$, which is similar to the estimated $\Delta$ in TTQ. \citet{hou_loss-aware_2018_r31} use the Newton proximal method to facilitate ternary quantization, the second-order information $\mathbf{d}$ is used as a scaling factor when calculating $\alpha$:
\begin{equation}
\small
\alpha=\frac{\|\mathbf{\hat{w}} \odot \mathbf{d} \odot \mathbf{w}\|_{1}}{\|\mathbf{\hat{w}} \odot \mathbf{d}\|_{1}}.
\end{equation}
A greedy search function $\texttt{find}(\cdot)$ is defined to help obtain $\alpha$. The $\Delta$ is assigned to $\frac{a}{2}$ (Corollary 3.1 in \cite{hou_loss-aware_2018_r31}). 
\citet{wang_two-step_2018_r27} use greedy search to find optimized $\alpha$. More specifically, during each iteration, it only updates one element of $\mathbf{\hat{w}}$ but fix all the other elements. It only needs to check $n\times 2^k$ possibilities, and the bit number $k$ is usually very small. \citet{hu_cluster_2019_r24} minimize a variant of Eq.~\eqref{eq:twn} to encourage weight values to get close to $\{0,\pm\alpha\}$:
\begin{equation}
\small
\underset{\alpha, z\in\{0,1\} }{\operatorname{argmin}}\left\|\mathbf{w}-\alpha \mathbf{\hat{w}}^{T} \mathbf{z}\right\|^{2},
\end{equation}
\begin{equation}
\small
\alpha=\frac{\mathbf{w} \mathbf{z}^{\top} \mathbf{\hat{w}}}{\mathbf{\hat{w}}^{\top} \mathbf{z} \mathbf{z}^{\top} \mathbf{\hat{w}}},
\end{equation}
where $\mathbf{z}$ denotes a sparsity indicator vector like a pruning mask. The author of \cite{hu_cluster_2019_r24} defines a greedy function $\texttt{H}(\cdot)$ to solve $\mathbf{z}$, i.e., finding a proper sparsity of $\mathbf{w}$ to obtain optimal $\alpha$. \citet{cavigelli_rpr_2020_r21} apply group-wise stochastic quantization and gradually increase the quantized portion. During initialization, the pre-trained weights are re-scaled by $\frac{w}{\alpha}$ in order to reduce the $L2$ distance between $\mathbf{w}$ and $\mathbf{\hat{w}}$ (Eq.~(5) in \cite{cavigelli_rpr_2020_r21}). As for finding the optimized $\alpha$, they apply grid search over 1,000 points spread uniformly over $[0, \operatorname{max}(\|\mathbf{w}\|_1)]$.
\citet{lei_memory_2020_r13} show a classical proximal mapping quantization. They quantize the SVM model to ternary values. The optimization process does not involve gradient descent (Eq.~\eqref{eq:ref13}).
Instead, it exhausts all of the possible values of $\{-1,0,1\}$ for each $w$ and selects the value that produces the minor loss.

In summary, although the direct projection is intuitive and convenient, the discrete weights will cause inaccurate gradients. Therefore, 
based on the proximal operator and the STE, many works propose different refining ideas to relieve the inaccurate gradient issue or get more accurate $\alpha$ and $\Delta$. However, there is no detailed analysis explaining the connection among $\alpha$, $\Delta$, and $\mathbf{w}$. Moreover, there was not any significant improvement in classification accuracy in recent years.

\subsection{Indirect Projection}
\label{indirect_projection}
Indirect projection can offset the disadvantage of direct projection based on a statement in the work of \citet{courbariaux2015binaryconnect}: \textit{\cite{muller2015rounding} and \cite{gupta2015deep} show that randomized or stochastic rounding can provide the unbiased discrete projection}. 

In the work of \citet{lin_neural_2016_r47}, they use a stochastic method in the forward pass and propose a constraint function $\bar{w}=\texttt{HardSigmoid}(w)$ \cite{courbariaux2015binaryconnect} to ensure $\bar{w}\in\{0,1\}$. In fact, $\bar{w}$ is a parameter of a random value generator. The stochastic quantization process is defined below:
\begin{equation}
\small
\begin{array}{l}P\left(\hat{w}=1\right)=\bar{w}; P\left(\hat{w}=0\right)=1-\bar{w}; \text{if}~\bar{w}>0 \\ P\left(\hat{w}=-1\right)=-\bar{w}; P\left(\hat{w}=0\right)=1+\bar{w}; \text{if}~\bar{w}\le 0\end{array}
\end{equation}
Nevertheless, stochastic quantization can only relieve the inaccurate gradient issue, and the sampling speed is slow when quantizing larger models. Although the experiment result of \cite{lin_neural_2016_r47} is better than the deterministic method, almost all of the following deterministic methods outperform the work of \cite{lin_neural_2016_r47}. In addition, the author of \cite{lin_neural_2016_r47} also shows the possibility of quantizing the gradient to 3 to 4 bits during backpropagation.
Unlike other works that directly update the full-precision copy of the weight, GXNOR-Net \cite{lei_deng_gxnor-net_2018_r30} projects the magnitude of gradients to transition probabilities. The stochastic ternary quantization is then controlled by such probability. 

The other similar work of \citet{shayer_learning_2018_r29} takes $\theta$ as the parameter of the multinomial distribution \texttt{M}() to generate discrete weight values $\hat{w}=\texttt{M}(\theta)$. To make the discrete values $\hat{w}$ continuous again, the author applies the following process:
\begin{equation}
\small
\begin{array}{l} \mu_{i j}=\mathbb{E}_{\theta_{i j}}\left[\hat{w}_{i j}\right] \text { and } \sigma_{i j}^{2}=\operatorname{Var}_{\theta_{i j}}\left[\hat{w}_{i j}\right], \\ m_{i}=\sum_{j} \mu_{i j} h_{j} \text { and } v_{i}^{2}=\sum_{j} \sigma_{i j}^{2} h_{j}^{2} \\ \epsilon \sim \mathcal{N}(0, I), \\ w=m+v\odot \epsilon,
\end{array}
\end{equation}
where $h$ denotes the input of a layer (Eq.~(4) in \cite{shayer_learning_2018_r29}). In other words, they apply the reparameterization trick \cite{kingma2015variational} to produce continuous weights by using the normally distributed noise $\epsilon$, mean $m_i$, variance $v_{i}^{2}$. In this way, the gradients can pass through the projection operation. 



Knowledge distillation (KD) can also be used to improve the performance of ternary quantization. \citet{hinton2015distilling} propose the knowledge distillation method to transfer knowledge from a larger neural network (teacher) to a smaller one (student). Applying KD to quantization is proposed by \citet{mishra_apprentice_2017_r42}; in their work, they compare different scenarios, for example, training the teacher and student network together or training the student with a pre-trained teacher network. In addition to comparing the teacher and student with the same structure, they also compare the scenarios with different structures. Their results show that the KD can significantly improve quantization performance, but applying KD with different network structures and schemes does not bring additional improvement.
\section{\textcolor{black}{Opportunities and Future Works}}
\label{sec:future}
The issue of ternary quantization is low training efficiency and poor accuracy. Most ternary quantization methods using large datasets are trained based on pre-trained models, which is more time-consuming. The lower accuracy is caused by using estimated gradients to update layer weights. Even with increased bit width, the estimation bias in STE still leads to a loss of accuracy since continuous space is essential when performing backpropagation. In addition, the commonly used speedup only works with 16-bit or 8-bit weights, and the hardware implementation for ternary inference has yet to be widely used. 
Besides, only a few works, such as TTQ et al., study the relationship between the model sparsity and quantization. From a codebook perspective, vector quantization (VQ) and ternary quantization are similar. Ternary can be seen as a special case of fixed codebooks, and more work can be done in this area. Ternary is mainly reflected in the optimization of memory consumption, and the advantage of VQ is mainly reflected in the optimization of hard disk space. The combination of the two can achieve the effect of can optimize both hard disk and memory consumption. There are more ways to design ternary quantization from a stochastic perspective and a generative model perspective.

Another problem is that most works use full-precision weights as initialization to obtain ternary weights. Intuitively, this may need to be revised. \citet{achterhold_variational_2018_r33.5} propose a training procedure for constrained full-precision models to obtain very accurate quantized models without training aware quantization. \citet{liu2023hyperspherical} mention that finding the best full-precision weights around ternary weights and then applying regular ternary quantization can boost the quantization performance. In other words, the current initialization method may need to be adjusted to improve the accuracy. In addition, general quantization methods may emerge in the future.
\section{\textcolor{black}{Conclusion}}
\label{sec:conclusion}

In this survey, we summarize the ternary quantization from the perspective of the projection function and optimization function. We first introduce a brief history of ternary quantization. For example, the origin of the first ternary quantization method and the ternary quantization methods before the introduction of STE. We introduce the projection functions, which include direct projecting, weight grouping, enlarging, stochastic rounding, temperature adjusting, and stochastic sampling. Our classifications provide a new
perspective different from conventional classifications, such as uniform, nonuniform, symmetric, and asymmetric quantization methods. Moreover, our classifications are applicable to higher-bit quantization works. We reveal the intrinsic relationship among the ternary quantization methods via proximal optimization. We show how ADMM can explain the widely applied alternating training scheme in ternary quantization. We also examine the connections among STE, dual-averaging, and proximal operators in ternary quantization. This survey summarizes the current research in ternary quantization and provides insight into higher-bit quantization works.

\newpage
\bibliographystyle{named}
\small
\bibliography{ijcai23}

\end{document}